\documentclass[sigconf]{acmart}
\usepackage{algorithm}
\usepackage{algorithmicx}
\usepackage{algpseudocode}
\usepackage{amsmath} % For math environments like equation
\usepackage{amsfonts} % For special math fonts

\usepackage{amssymb} % For additional math symbols
\usepackage{array} 
\usepackage{booktabs}     
\usepackage{caption}
\usepackage{siunitx}
\usepackage{float}
\usepackage{graphicx}
\usepackage{placeins}
\usepackage{multirow}
\usepackage{makecell}      % For adjustable cell gap with \makegapedcells  
\usepackage{subcaption}
\usepackage[utf8]{inputenc}
\usepackage{hyperref} % For hyperlinks in citations
\usepackage{tikz}
\usetikzlibrary{positioning}
\floatname{algorithm}{Algorithm}

\usepackage{titlesec}
\titlespacing*{\section}{0pt}{*0}{*0} 
\titlespacing*{\subsection}{0pt}{*0}{*0} 
\copyrightyear{2024}
\acmYear{2024}
\acmDOI{XXXXXXX.XXXXXXX}
\begin{document}
%%%%%%%%%%%%%%%%%%%%%%%%%%%%%%%%%%%%%%%%%%%%%%%%%%%%%%%%%%%%%%%%%%%%%%%%%%%
\title{DynamicRouteGPT: A Real-Time Multi-Vehicle Dynamic Navigation Framework Based on Large Language Models}
%%%%%%%%%%%%%%%%%%%%%%%%%%%%%%%%%%%%%%%%%%%%%%%%%%%%%%%%%%%%%%%%%%%%%%%%%%%%%%%%
\author{Ziai Zhou}
\authornote{This author contributed the most to this research.}
\affiliation{%
  \institution{Shandong University}
  \city{Qingdao}
  \country{China}
}
\email{202000210104@mail.sdu.edu.cn}
\author{Bin Zhou}
\affiliation{%
  \institution{Shandong University}
  \city{Qingdao}
  \country{China}
  }
\email{binzhou@sdu.edu.cn}
\author{Hao Liu}
\authornote{Corresponding author.}
\affiliation{%
  \institution{The Hong Kong University of
Science and Technology (Guangzhou)}
  \city{Guangzhou, Guangdong}
  \country{China}
}
\email{liuh@ust.hk}
\renewcommand{\shortauthors}{Ziai Zhou}
\begin{abstract}
\textbf{Draft Version: This is a preliminary draft of the paper and is currently under review. The content may be updated or revised in future versions.}
Real-time dynamic path planning in complex traffic environments presents significant challenges, such as varying traffic volumes and traffic signal wait times. While traditional static routing algorithms like Dijkstra and A* have been successful in computing the shortest paths, they often fail to effectively address the dynamic nature of traffic conditions. Recent Reinforcement Learning (RL) approaches have made some progress in dynamic navigation; however, most existing models focus solely on local optima, neglecting global optimality and risking dead-ends or network boundary issues in practical applications.This paper proposes a novel approach based on causal inference, aimed at achieving real-time dynamic path planning for multiple vehicles in real-world road networks, balancing both global and local optimality. We first utilize the static Dijkstra algorithm to compute a globally optimal path as a baseline. Then, a distributed control strategy directs vehicles along the baseline path. Upon reaching intersections, DynamicRouteGPT performs real-time decision-making for local path selection, considering real-time traffic information, driving preferences, and unexpected events to optimize dynamic path choices at each intersection.The DynamicRouteGPT framework integrates multiple techniques, including Markov chains, Bayesian inference, and modern large-scale pretrained language models such as Llama3 8B, to provide an efficient and reliable path planning solution. This framework not only processes real-time traffic data but also dynamically adjusts path planning to accommodate various traffic scenarios and driver preferences. A key innovation of our method lies in constructing causal graphs for counterfactual reasoning, optimizing path decisions. Our model requires no pre-training, boasts broad applicability, and can handle various types of road networks. Experimental results demonstrate that this method achieves state-of-the-art (SOTA) performance in real-time dynamic path planning for multiple vehicles while offering explainability in path selection. This approach not only effectively selects the optimal path but also provides justifications for the choices made, offering a novel and efficient solution for dynamic navigation in complex traffic environments.
\end{abstract}
%%%%%%%%%%%%%%%%%%%%%%%%%%%%%%%%%%%%%%%%%%%%%%%%%%%%%%%%%%%%%%%%%%%%%%%%%%
%%%%%%%%%%%%%%%%%%%%%%%%%%%%%%%%%%%%%%%%%%%%%%%%%%%%%%%%%%%%%%%%%%%%%%%%%%%%%%%%%
%%%%%%%%%%%%%%%%%%%%%%%%%%%%%%%%%%%%%%%%%%%%%%%%%%%%%%%%%%%%%%%%%%%%%%%%%%%%%%%%
\maketitle
%%%%%%%%%%%%%%%%%%%%%%%%%%%%%%%%%%%%%%%%%%%%%%%%%%%%%%%%%%%%%%%%%%%%%%%%%%%%%%%%%%%%
\section{Introduction}
\begin{figure*}[htbp]
    \centering
    \includegraphics[width=\textwidth]{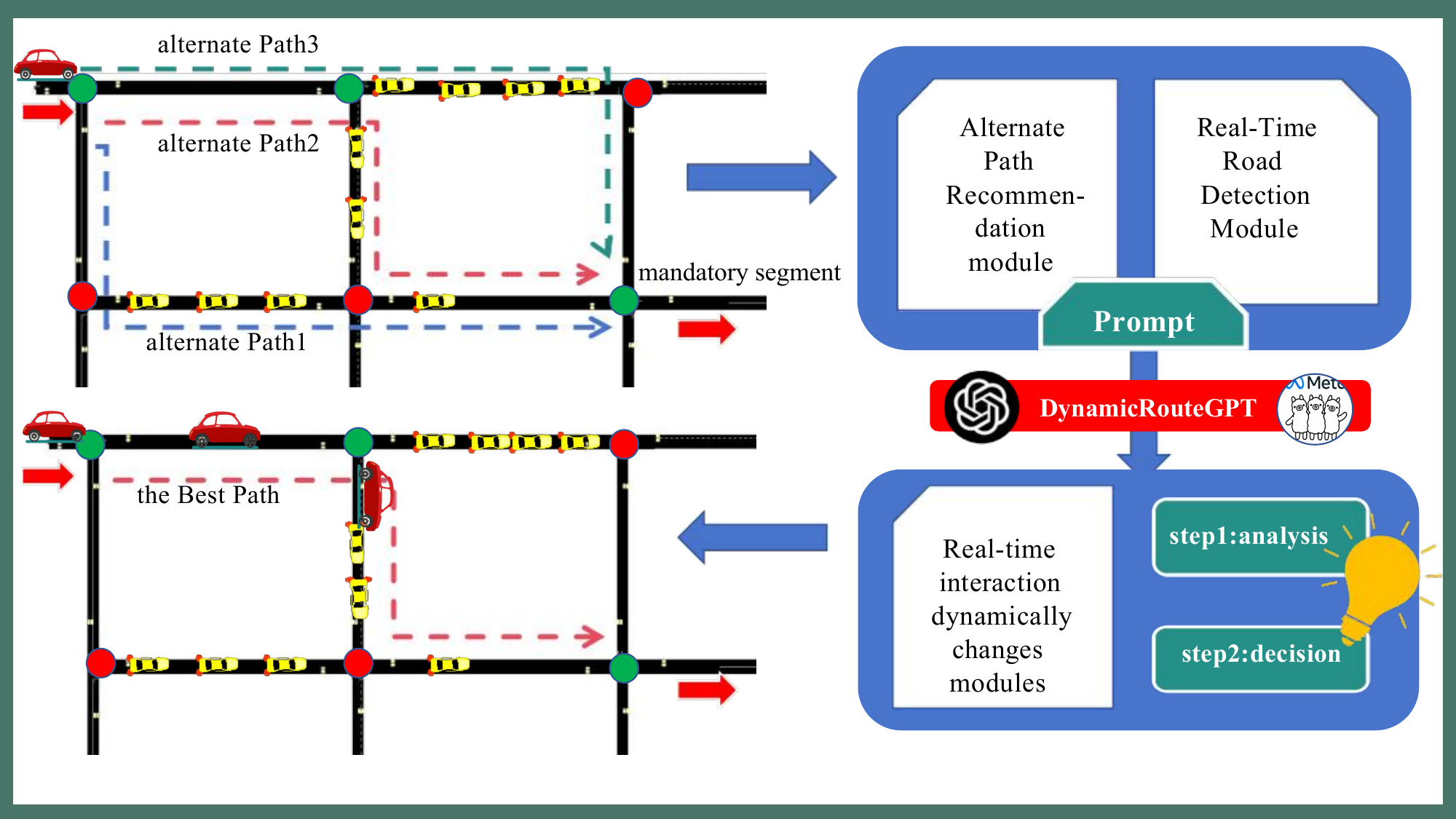}
    \caption{An overview of the DynamicRouteGPT framework, illustrating the main modules, data flow, and interactions in the path planning process.}
    \label{fig:framework}
\end{figure*}

Real-time multi-vehicle dynamic path planning in modern intelligent transportation systems presents a highly challenging problem. As urban road networks continue to expand and traffic flows become increasingly complex, traditional path planning methods often struggle to perform effectively in dynamic environments\cite{r4}. Specifically, fluctuations in traffic volume, periodic changes in traffic signals, and frequent traffic incidents necessitate a path planning system with high adaptability and responsiveness. Furthermore, the variability in road conditions, such as traffic congestion, road damage, and adverse weather, places additional demands on the real-time nature of path planning\cite{r7,r22}. Path selection for multiple vehicles not only needs to consider the optimal route for individual vehicles but also must coordinate their interactions to avoid traffic jams or potential accidents. This coordination between global and local optimization is one of the key challenges in achieving effective path planning.\newline
With advancements in autonomous driving technology, issues related to vehicle trajectory control and navigation have gained increasing attention. However, existing approaches still face several limitations in practical applications. Classical path planning algorithms, such as Dijkstra  and A* \cite{r25}, are effective at calculating the shortest path under static conditions but are significantly constrained in dynamic environments due to their inability to adapt to real-time changes in road conditions. Recently, dynamic navigation models based on reinforcement learning (RL), such as XRouting \cite{r3} and DQN navigation \cite{r23,r12}, have attracted widespread attention for attempting to address these challenges. These models seek to achieve dynamic path optimization in real-time environments by framing the path planning task as a series of decision-making problems \cite{r26,r12,r27}. Specifically, they formulate the task as an RL problem where the vehicle learns to make decisions—such as turning left, right, or going straight at upcoming intersections—based on its current position, destination information, and real-time traffic data. However, these approaches often focus too narrowly on local optima, neglecting the overall effectiveness of the global path. This can result in vehicles getting trapped in dead-ends or straying beyond network boundaries, ultimately preventing them from reaching their destinations\cite{r4}.\newline
To address these issues, this paper proposes a novel dynamic path planning framework based on large language models (LLMs), termed DynamicRouteGPT. DynamicRouteGPT is fine-tuned specifically for real-time multi-vehicle path planning, integrating driver preferences and real-time road information to dynamically adjust the routes of multiple vehicles in response to complex scenarios, such as forced detours and unexpected road closures. The innovation of this framework lies in its combination of global path optimization and local path dynamic adjustment, achieving a balance between global and local optima by predicting road conditions and traffic flow changes.
To better illustrate the overall structure of this work and the proposed framework, Figure \ref{fig:framework} presents an overview of the DynamicRouteGPT framework. This figure details the main modules, data flow, and their interactions within the path planning process. Through this visualization, readers can gain a clear understanding of the framework's workflow and how it integrates various technologies to optimize multi-vehicle path planning.\newline
In experiments, DynamicRouteGPT demonstrated significant advantages in multi-vehicle control and overall time optimization, surpassing the current state-of-the-art (SOTA) methods. By leveraging the strong predictive capabilities of LLMs, DynamicRouteGPT not only improves the efficiency and accuracy of path planning but also significantly reduces computational costs, offering a promising new solution for dynamic path planning in intelligent transportation systems. Compared to traditional methods, DynamicRouteGPT exhibits greater adaptability in handling dynamic changes in complex road networks, enabling real-time optimization of multi-vehicle path selection while minimizing travel time and avoiding traffic congestion.\newline
The main contributions of this paper can be summarized as follows:
\begin{itemize}
    \item We propose DynamicRouteGPT, a novel dynamic path planning framework based on LLMs, specifically designed to address the complex challenges of real-time multi-vehicle navigation.
    \item We introduce a method that combines global and local optima, successfully resolving the conflict between local and global optimization in traditional path planning, ensuring that vehicles can reach their destinations smoothly in dynamic environments.
    \item We establish a new baseline for dynamic path planning under real-time traffic conditions, providing a reference for evaluating the performance of path planning algorithms in diverse road network environments.
    \item Through experimental validation, DynamicRouteGPT outperforms existing SOTA methods in multi-vehicle control and overall time optimization, demonstrating its effectiveness and potential in real-world traffic scenarios.
    \item This paper provides a new perspective on multi-vehicle dynamic path planning in intelligent transportation systems, exploring the application possibilities of LLMs in this domain, thereby laying the foundation for future research and applications.
\end{itemize}
In summary, DynamicRouteGPT applies advanced language model technology to dynamic path planning, achieving both precision and real-time responsiveness in path selection, while offering an innovative solution to the complex problems encountered in real-world traffic scenarios. This research not only holds significant theoretical value but also provides a new technical pathway for the practical application of intelligent transportation systems.\newline
%%%%%%%%%%%%%%%%%%%%%%%%%%%%%%%%%%%%%%%%%%%%%%%%%%%%%%%%%%%%%%%%%%%%%%%%%%%%%%%%%
\section{Preliminaries}
In this section, we first introduce the key concepts involved in the task of Dynamic Navigation control for multiple vehicles under real-time road conditions.
%%%%%%%%%%%%%%%%%%%%
\subsection{Road Network}

A road network is a system of interconnected roads that forms a mesh-like distribution within a specific region\cite{r1}. For simulation purposes, we employ SUMO (Simulation of Urban MObility)\cite{r31}, where the road network in SUMO is represented as a directed graph. In SUMO, nodes typically represent intersections, while edges represent roads or streets, which are unidirectional. Specifically, a SUMO network includes the following information: each edge consists of a collection of lanes, with each lane having an ID, location, shape, and speed limit. Intersections are equipped with traffic signal logic, and junctions manage right-of-way regulations. Connections between lanes are represented by nodes at intersections.
%%%%%%%%%%%%%%%%%%%%
\subsection{Dynamic Vehicle Navigation}

The goal of vehicle path planning is to find the route that minimizes the cost for a vehicle to reach its destination. Classical routing solutions, such as the Dijkstra algorithm  and A* \cite{r25}, greedily compute the shortest path between the starting node and the target node. While these static routing algorithms are widely adopted in practice, they do not account for the dynamic changes in road conditions in real time. Recently, several works have proposed learning-based models for real-time vehicle navigation, where the navigation task is framed as a series of rerouting decisions \cite{r26,r12,r27}. These models formulate the task as a reinforcement learning (RL) problem, where the vehicle learns to turn left, right, or go straight at upcoming intersections based on its observations, including current location, destination information, and real-time traffic conditions. However, this single-path decision-making approach is prone to getting trapped in local optima, leading to situations where vehicles become stuck in dead ends or deviate from the road network boundaries, failing to reach their destinations \cite{r4}.
%%%%%%%%%%%%%%%%%%%%
\subsection{Multi-Vehicle Control}

Multi-vehicle control (MVC) represents a significant research area in robotics, autonomous systems, and transportation systems, focusing on the coordination and control of multiple vehicles to efficiently execute collective tasks. The core of this field lies in developing algorithms and strategies to ensure that a group of vehicles can maintain specific spatial positions relative to each other while collaboratively completing assigned tasks. This process involves complex strategy coordination, including task allocation, communication protocol design, and route planning and obstacle avoidance strategies. Through effective communication mechanisms, vehicles can exchange information to avoid potential conflicts on their paths, thereby reducing the risk of traffic accidents. Additionally, each vehicle must independently plan its trajectory to ensure it does not collide with other vehicles or environmental obstacles, ultimately optimizing the total time for the entire fleet. Models optimized with LoRA can more effectively incorporate Bayesian reasoning to select the optimal path and provide explanations.
%%%%%%%%%%%%%%%%%%%%%%%%%%%%%%%%%%%%%%%%%%%%%%%%%%%%%%%%%%%%%%%%%%%%%%%%%%%%
\section{Related Work}

In this section, we review the literature on the use of large language models (LLMs) in traffic signal control, end-to-end vehicle control, and decision-making applications. We begin by exploring LLMs in traffic signal control agents, followed by their role in autonomous driving systems. Lastly, we discuss LLMs in broader decision-making contexts, highlighting their relevance to intelligent transportation systems. This review provides context for our research and positions it within the existing landscape of LLM applications.

\subsection{LLMs for Traffic Signal Control Agents}

Traffic signal control is a critical component of urban infrastructure management. In recent years, several studies have explored the use of Large Language Models (LLMs) to enhance the decision-making capabilities of traffic signal control agents. \cite{r1} introduced LLMlight, a method that leverages LLMs to generate introspective suggestions for optimizing traffic signal control strategies. These suggestions are derived from the agent's past experiences, expert demonstrations, and generalization across different traffic scenarios. Unlike traditional reinforcement learning approaches, LLMlight does not require parameter fine-tuning but instead adapts the prompts provided to the agent, enabling it to effectively adjust to new situations. This method demonstrated significant improvements in traffic flow and effectively reduced congestion in simulation experiments.
%%%%%%%%%%%%%%%%%%%%
\subsection{LLMs for End-to-End Vehicle Control Tasks}

Multimodal Large Language Models (MLLMs) have shown exceptional performance in processing and reasoning about non-textual data such as images and videos, emerging as a promising area of research. These models have been applied to autonomous driving, leading to the development of interpretable end-to-end systems like DriveGPT4 \cite{r17,r18}. DriveGPT4 can process multi-frame video inputs and textual queries to interpret vehicle actions and provide reasoning. The system is capable of predicting low-level vehicle control signals in an end-to-end manner and has outperformed existing state-of-the-art models such as GPT4-V on the BDD-X dataset \cite{r21}.
%%%%%%%%%%%%%%%%%%%%
\subsection{LLMs in Decision-Making Applications}
The emergence of LLMs has significantly impacted the field of natural language processing, achieving remarkable results across various tasks, including decision-making. In the context of decision-making, LLMs have been employed to enhance the performance of agents in diverse domains. For instance, \cite{r16} explored the use of LLMs in decision support systems, where these models are used to generate concise and valuable suggestions to optimize decision-making processes. This approach has been tested in various settings, including traffic signal control and autonomous driving, showing significant improvements in both few-shot and zero-shot learning scenarios.

In summary, the application of LLMs in decision-making for traffic signal control and autonomous driving has shown great potential. Techniques like LLMlight and DriveGPT \cite{r17,r18} demonstrate the capability of LLMs to enhance the adaptability and efficiency of these systems. However, further research is needed to fully exploit the capabilities of LLMs in these areas and to address the remaining challenges.

%%%%%%%%%%%%%%%%%%%%%%%%%%%%%%%%%%%%%%%%%%%%%%%%%%%%%%%%%%%%%%%%%%%%%%%%
\section{Methodology}

To address the challenges of dynamic traffic management and provide optimal path planning solutions for multiple vehicles, we model the problem using a mathematical framework. The objective of our research is to develop a system capable of dynamically adjusting routes in real-time, ensuring both global and local optimality while responding to unforeseen events such as road closures and temporary detours. Additionally, the model aims to incorporate driver preferences, such as specific routes to be taken or paths to be avoided, to enhance the accuracy and efficiency of the path planning process.

We represent the road network as a graph \( G = (V, E) \), where \( V \) is the set of nodes representing intersections, and \( E \) is the set of edges representing road segments. Each edge \( e \in E \) is associated with multiple dynamic attributes, such as length \( l(e) \), speed limit \( v_{\text{max}}(e) \), current average speed \( v_{\text{avg}}(e) \), and traffic signal information. The objective of the problem is to find a path \( P = \{e_1, e_2, \dots, e_n\} \) that minimizes the total travel time, considering real-time and predictive traffic conditions.
%%%%%%%%%%%%%%%%%%%%%%%%%%%%%%%%%%%%%%
\subsection{DynamicRouteGPT Framework}

We propose the DynamicRouteGPT framework, which comprises four core modules, as illustrated in the flowchart (Figure \ref{fig:framework}). These modules include the Real-Time Information Retrieval Module, the Alternative Path Generation Module, the GPT Decision Module, and the Dynamic Path Adjustment Module. Together, these modules form the overall architecture of the system, enabling dynamic, efficient path planning and real-time adjustments.

In this section, we provide a detailed explanation of the design concepts, innovations, and integration of each module within the overall system. To aid in understanding the implementation of these modules, the pseudocode provided in Algorithm  outlines the core logic of the framework, specifically describing the interactions and processes between the different components of the framework.

\begin{tikzpicture}[node distance=2.5cm, auto, scale=0.8, transform shape]

    % Define the nodes
    \node [draw, rectangle, rounded corners, text width=3.5cm, align=center] (module1) {
        \textbf{01 Alternate Path\\Recommendation Module} \\
        \small On the basis of determining the global optimal path, local alternative paths are provided.
    };
    \node [draw, rectangle, rounded corners, text width=3.5cm, align=center, right=of module1] (module2) {
        \textbf{02 Necessary Information\\Retrieval Module} \\
        \small Provide real-time information on the alternate routes available for the current road for model decision-making.
    };
    \node [draw, rectangle, rounded corners, text width=3.5cm, align=center, below=of module2] (module3) {
        \textbf{03 Interaction Between DriverRouteGPT And Vehicle} \\
        \small The LLMs select the route based on the provided alternative routes.
    };
    \node [draw, rectangle, rounded corners, text width=3.5cm, align=center, left=of module3] (module4) {
        \textbf{04 Vehicle Control Module} \\
        \small Integrate the route selection made by LLMs with real-time traffic data to dynamically adjust the vehicle's route.
    };

    % Draw the arrows
    \draw [->, thick] (module1) -- (module2);
    \draw [->, thick] (module2) -- (module3);
    \draw [->, thick] (module3) -- (module4);
    \draw [->, thick] (module4) -- (module1);

\end{tikzpicture}

%%%%%%%
\subsubsection{Real-Time Information Acquisition Module}  

The purpose of the Real-Time Information Acquisition Module is to obtain real-time vehicle location data within the road network, as well as relevant traffic information, to optimize routing under dynamic conditions. To accurately model dynamic traffic states, we employ a Markov chain model \cite{r33,r34} to characterize the state transition process of road segments and predict travel times.

Let the road network be represented by a graph \( G = (V, E) \), where \( V \) denotes the set of nodes (intersections) and \( E \) denotes the set of edges (road segments). Each edge \( e \in E \) represents a state, and the attributes associated with each edge include its length \( l(e) \), speed limit \( v_{\text{max}}(e) \), and current average speed \( v_{\text{avg}}(e) \).

We define the state space \( S \) as the set of all road states, i.e., \( S = \{s_1, s_2, \dots, s_n\} \), where \( s_i \) represents the state of edge \( e_i \). The state transition probability \( P(s_{i+1} | s_i) \) denotes the probability of a vehicle transitioning from state \( s_i \) to state \( s_{i+1} \).

Considering traffic signals and other factors that may influence traffic flow, we define the state transition probability as:
\[
P(s_{i+1} | s_i) = f\left( \frac{v_{\text{avg}}(e_i)}{v_{\text{max}}(e_i)}, \frac{\text{signal wait time}(e_i)}{\text{signal cycle}(e_i)}, \text{traffic density}(e_i) \right)
\]
where the function \( f \) is trained on historical and real-time data, incorporating the effects of multiple factors.

The travel time \( t(e) \) for each edge \( e \) can be expressed as:
\[
t(e) = \frac{l(e)}{v_{\text{avg}}(e)} + \text{signal wait time}(e)
\]

By utilizing the Markov chain model, we can predict the transition times between different states of the vehicle, thereby providing essential prior information for the path decision-making in DynamicRouteGPT.
%%%%%%%%%
\subsubsection{Alternative Path Generation Module}

The purpose of the Alternative Path Generation Module is to generate multiple candidate paths to ensure local optimality while maintaining global optimality. Existing reinforcement learning-based navigation models typically focus on action selection at local intersections, which often leads to local optima and neglects the global optimal path.

To address this issue, we first utilize the Dijkstra algorithm to compute the global optimal path \( P_{\text{global}} \) under a static road network. For each critical segment (a mandatory segment within the global path), we then calculate multiple shortest paths \( \{P_1, P_2, P_3\} \) from the current segment to the next or the subsequent critical segment, based on real-time traffic conditions. These paths serve as alternative options, ensuring that the vehicle can make optimal decisions while adhering to the global optimal path during actual navigation.

%%%%%%%%%%
\subsubsection{GPT Decision-Making Module}

The GPT Decision-Making Module leverages the state transition probabilities provided by the Real-Time Information Acquisition Module and the candidate paths generated by the Alternative Path Generation Module to perform path selection through Bayesian inference. Specifically, this module applies Bayes' theorem to update the probability distribution of each path, thereby facilitating the selection of the optimal route:
\[
P(P_i \mid \text{real-time information}) = \frac{P(\text{real-time information} \mid P_i) \cdot P(P_i)}{P(\text{real-time information})}
\]
where \( P(P_i) \) denotes the prior probability of selecting path \( P_i \), derived from historical data, and \( P(\text{real-time information} \mid P_i) \) represents the likelihood of observing the current real-time traffic data given that path \( P_i \) is chosen. The posterior probability \( P(P_i \mid \text{real-time information}) \) indicates the likelihood that path \( P_i \) is the optimal route under the current real-time conditions.

The workflow of the GPT Decision-Making Module is as follows:
\begin{enumerate}
    \item \textbf{Input Processing}: The model receives input data, including vehicle status, traffic conditions, and candidate paths.
    \item \textbf{Path Probability Calculation}: The posterior probability \( P(P_i | \text{real-time information}) \) of each candidate path is calculated using Bayes' theorem.
    \item \textbf{Path Selection}: The path with the highest posterior probability is selected, and the rationale for this selection is generated.
    \item \textbf{Output and Adjustment}: Based on the selected path, the system dynamically adjusts the vehicle's navigation instructions.
\end{enumerate}

Through this process, DynamicRouteGPT is capable of providing efficient and precise path planning in complex and dynamic traffic environments, while dynamically adapting to new traffic conditions. The Bayesian inference framework enables the system to continuously update its path selection as new information becomes available, ensuring that the chosen route is contextually relevant and responsive to real-time traffic conditions.

%%%%%%%%%%
\subsubsection{Dynamic Path Adjustment Module}

The Dynamic Path Adjustment Module is responsible for implementing the decisions made by the GPT Decision-Making Module in the actual driving process and dynamically adjusting the route based on real-time changes. This module continuously retrieves the latest traffic data from the Real-Time Information Acquisition Module and, based on the output from the GPT module, updates the vehicle’s route in real time to ensure that it remains on the optimal path.

This process is analogous to feedback control in control theory, wherein system parameters are continuously adjusted to achieve optimal performance. By iteratively refining the vehicle's path in response to dynamic traffic conditions, the Dynamic Path Adjustment Module plays a critical role in maintaining the efficiency and effectiveness of the overall path planning system, ensuring that the vehicle consistently follows the most efficient route under varying traffic scenarios.
%%%%%%%%%%%%%%%%%%%%%%%%%%%%%%%%%%
\subsection{DynamicRouteGPT}

To achieve efficient and precise path planning, we have employed the Llama3 8B model \cite{r35}, a large-scale pre-trained language model capable of handling complex input data. The Llama3 8B model, trained on a diverse and extensive dataset, exhibits exceptional contextual understanding and reasoning capabilities, enabling accurate predictions and decision-making in complex traffic scenarios.

To fully harness the potential of the Llama3 8B model, we have integrated it with the LLAMA-Factory inference framework. LLAMA-Factory \cite{r36} is a powerful and user-friendly platform designed for efficient training and fine-tuning of large language models, supporting their deployment across various traffic scenarios. Through LLAMA-Factory, we can swiftly adapt the Llama3 8B model to different traffic contexts and dynamically adjust its inference process to respond to rapidly changing traffic conditions. The modular design of this framework allows the model to flexibly adjust to the demands of different scenarios, thereby maintaining efficient path selection in complex traffic environments.

During the fine-tuning process, we employed the BAdam optimization algorithm \cite{r37} to specifically adjust the Llama3 8B model. BAdam, a variant of the Adam optimization algorithm, adaptively adjusts the learning rate to better accommodate the dynamic changes in traffic scenarios. The BAdam algorithm updates the model parameters using the following equations:
\[
\theta_{t+1} = \theta_t - \alpha \cdot \frac{\hat{m}_t}{\sqrt{\hat{v}_t} + \epsilon}
\]
where \(\theta_t\) represents the model parameters at time step \(t\), \(\alpha\) is the learning rate, and \(\hat{m}_t\) and \(\hat{v}_t\) are the bias-corrected estimates of the first and second moments, respectively, computed as:
\[
\hat{m}_t = \frac{m_t}{1 - \beta_1^t}, \quad m_t = \beta_1 m_{t-1} + (1 - \beta_1) g_t
\]
\[
\hat{v}_t = \frac{v_t}{1 - \beta_2^t}, \quad v_t = \beta_2 v_{t-1} + (1 - \beta_2) g_t^2
\]
Here, \(g_t\) denotes the gradient, \(\beta_1\) and \(\beta_2\) are hyperparameters typically set to \(\beta_1 = 0.9\) and \(\beta_2 = 0.999\), and \(\epsilon\) is a small constant to prevent division by zero.

By optimizing with BAdam, the model maintains its efficiency and accuracy in path planning even in the presence of challenges such as traffic congestion, road closures, and unforeseen events. This optimization strategy enables the model to perform robust navigation decisions, ensuring reliable performance in complex environments.

Additionally, to further enhance the model's adaptability and specificity, we applied Low-Rank Adaptation (LoRA) to fine-tune the Llama3 8B model. LoRA introduces low-rank matrices \(W = A \cdot B\) to fine-tune the pre-trained model's parameters, where \(A\) and \(B\) are low-rank matrices. The parameter update is performed as follows:
\[
\theta_{t+1} = \theta_t + \Delta W
\]
\[
\Delta W = \alpha \cdot (A \cdot B)
\]
where \(\Delta W\) represents the parameter adjustment after low-rank adaptation, and \(\alpha\) is the learning rate. LoRA enables the model to enhance its expressiveness and adaptability without significantly increasing computational costs, allowing it to better address the specific challenges of path planning tasks. This fine-tuning method not only improves the model's efficiency in handling real-time traffic data but also enhances its generalization capability and stability in practical applications.

Through the integration of these technologies, DynamicRouteGPT provides efficient and accurate path planning services in complex and dynamic traffic environments. The Llama3 8B model offers a robust foundation for inference, the LLAMA-Factory framework ensures flexibility and efficiency in the inference process, and the BAdam optimization algorithm combined with LoRA fine-tuning allows the model to effectively tackle the challenges of real-world path planning, ensuring optimal navigation outcomes across various traffic conditions.
%%%%%%%%%%%%%%%%%%%%%%%%%%%%%%%%%%
\subsection{Overall Design Motivation and Integration}
The DynamicRouteGPT framework is meticulously structured to ensure that the output of each module is seamlessly passed to the next, maintaining coherence across the entire system. The integration of the framework is designed to guarantee that local path adjustments do not compromise global optimization, while ensuring that the system retains sufficient robustness when responding to unexpected traffic conditions. By employing Markov chains to model state transitions and utilizing Bayesian inference for decision-making, the framework strikes a balance between theoretical rigor and practical applicability.

In summary, the DynamicRouteGPT framework represents a significant advancement in real-time vehicle navigation by proposing an innovative approach that combines traditional optimization techniques with modern AI-driven decision-making. This integrated approach enables the framework to deliver efficient and adaptive path planning solutions that cater to the demands of both autonomous and human-driven vehicles in complex urban environments.

DynamicRouteGPT achieves a groundbreaking integration of Bayesian inference with a GPT-based decision-making model, marking an innovative step forward in dynamic, multi-vehicle path planning. The Real-Time Information Acquisition Module ensures data accuracy, the Alternative Path Generation Module balances global and local considerations, and the GPT Decision-Making Module, powered by Bayesian reasoning, provides a scientifically grounded basis for path selection. Finally, the Dynamic Path Adjustment Module translates these theoretical advancements into practical applications. This framework effectively addresses the uncertainties inherent in complex traffic environments, offering a novel technological pathway for future intelligent transportation systems.
%%%%%%%%%%%%%%%%%%%%%%%%%%%%%%%%%%%%%%%%%%%%%%%%%%%%
\begin{algorithm}[htbp]
    \caption{Dynamic Route Planning and Adjustment using SUMO and GPT-based Decision Module}
    \begin{algorithmic}[1] % 每行显示行号
        \Require Network configuration: $net\_file$, Route configuration: $rou\_file$, Simulation settings: $cfg\_file$
        \Ensure Efficient and adaptive route planning and real-time adjustments
        \State \textbf{Initialize} the SUMO simulation environment with GUI and load network, route, and configuration files.
        \State \textbf{Start} the SUMO simulation using \texttt{traci.start(sumoCmd)}.
        \State $step \gets 0$
        \State $passed\_edges \gets []$, $passed\_nodes \gets []$
        \State $shortest\_path, shortest\_edge\_list \gets \texttt{get\_shortest\_path}(start\_edge, end\_edge)$
        \For{each vehicle $v_i$ in the vehicle set $V$}
            \If{a valid $shortest\_path$ is found}
                \State \textbf{Assign} $route\_id$ based on $shortest\_route$
                \State \textbf{Add} this route to the vehicle using \texttt{traci.route.add(route\_id, shortest\_edge\_list)}
                \State \textbf{Deploy} the vehicle into the simulation using \texttt{traci.vehicle.add(v\_id, route\_id)}
            \EndIf
        \EndFor
        \While{the simulation is running and vehicles are still active}
            \State \textbf{Advance} the simulation by one step using \texttt{traci.simulationStep()}
            \State $step \gets step + 1$
            \For{each active vehicle $v_i$ in the simulation}
                \State \textbf{Identify} the current edge $current\_edge$ from vehicle's lane ID
                \If{$current\_edge$ is not at an intersection}
                    \State \textbf{Determine} the next node $next\_node$ the vehicle is heading towards
                    \If{$next\_node$ has been visited before}
                        \State \textbf{Calculate} $k$ alternative paths using \texttt{find\_k\_shortest\_paths}(3 paths)
                        \State \textbf{Estimate} travel times and count traffic lights for each path
                        \State \textbf{Generate} a decision prompt for the GPT module based on current data
                        \State \textbf{Select} the optimal path $chosen\_path$ and obtain rationale using \texttt{GPT\_decision}(prompt)
                        \State \textbf{Update} the vehicle's route to follow $chosen\_path$
                    \Else
                        \State \textbf{Record} the $current\_edge$ and $next\_node$ as visited
                    \EndIf
                \EndIf
            \EndFor
        \EndWhile 
        \State \textbf{Terminate} the simulation and clean up using \texttt{traci.close()}
    \end{algorithmic}
\end{algorithm}

%%%%%%%%%%%%%%%%%%%%%%%%%%%%%%%%%%%%%%%%%%%%%%%%%%%%%%%%%%%%%%%%%%%%%%%%%
\section{Experiments}

In this section, we demonstrate the performance of the DynamicRouteGPT framework on dynamic route planning tasks and conduct comparative analysis with several representative baseline methods. The experiments cover areas such as traffic signal control, dynamic navigation, and joint control tasks. We also validate the effectiveness of each module through ablation studies and analyze the model's generalization ability and performance across different datasets. All experiments were conducted using the SUMO simulation platform, and the results were quantitatively analyzed using multiple evaluation metrics.
%%%%%%%%%%%%%%%%%%%%%%%%%%%%%%%%%%%%%%
\subsection{Simulation}

During the simulation process, we employed SUMO (Simulation of Urban MObility) to model real-world traffic dynamics, including vehicle movements and traffic signal phase changes. For human-driven vehicles (HVs), route planning data were directly obtained from the road network and route files, and the vehicles were controlled through SUMO. We utilized SUMO's built-in output functions to export the relevant simulation data. For autonomous vehicles (AVs), their driving patterns were determined by randomly sampling from valid routes based on the total number of vehicles entering the network during the simulation. 
The microscopic behavior of all vehicles was automatically modeled using SUMO's default built-in car-following model—the Intelligent Driver Model (IDM). This model adaptively controls acceleration based on each vehicle's speed and the distance to the vehicle ahead.

%%%%%%%%%%%%%%%%%%%%%%%%%%%%%%%%%%%%%%
\subsection{Dataset Description}

\begin{figure}[!t]
  \centering
    \begin{subfigure}{0.23\textwidth}
    \includegraphics[width=\linewidth]{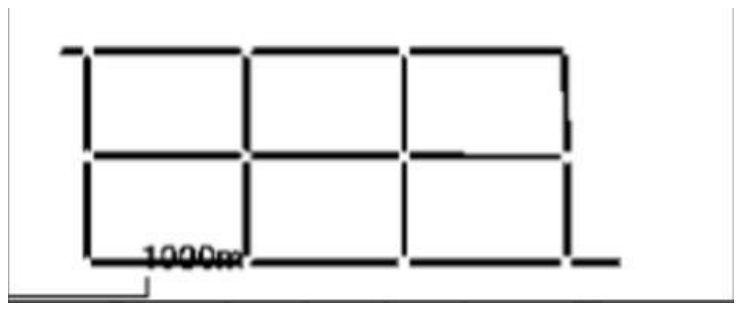}
    \caption{Manhattan}
    \label{fig:manhattan}
  \end{subfigure}
  \hfill
  \begin{subfigure}{0.23\textwidth}
    \includegraphics[width=\linewidth]{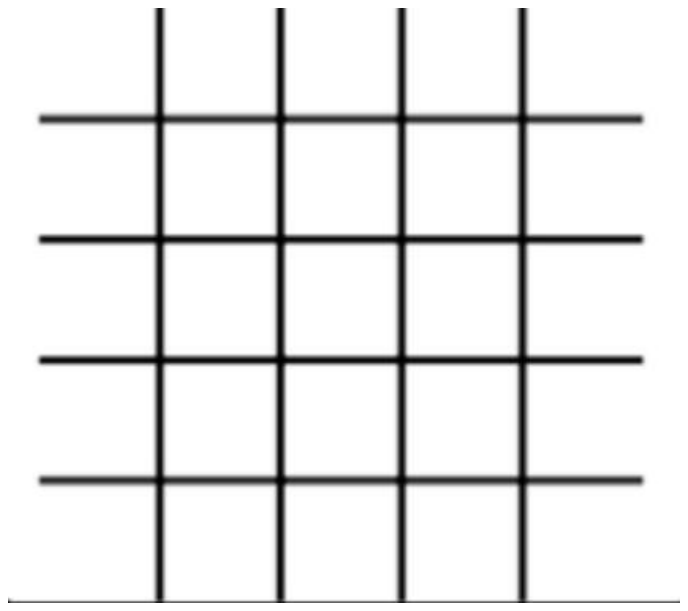}
    \caption{Grid 4x4}
    \label{fig:grid}
  \end{subfigure}
  \hfill
  \begin{subfigure}{0.23\textwidth}
    \includegraphics[width=\linewidth]{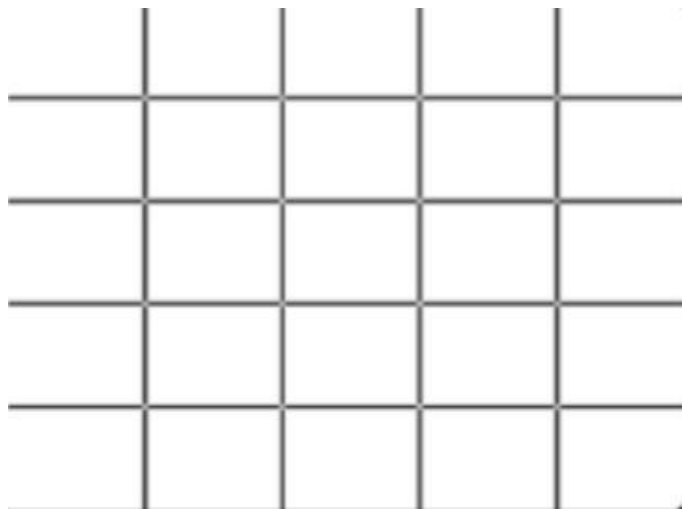}
    \caption{Hangzhou 4x4}
    \label{fig:hangzhou}
  \end{subfigure}
  \hfill
  \begin{subfigure}{0.23\textwidth}
    \includegraphics[width=\linewidth]{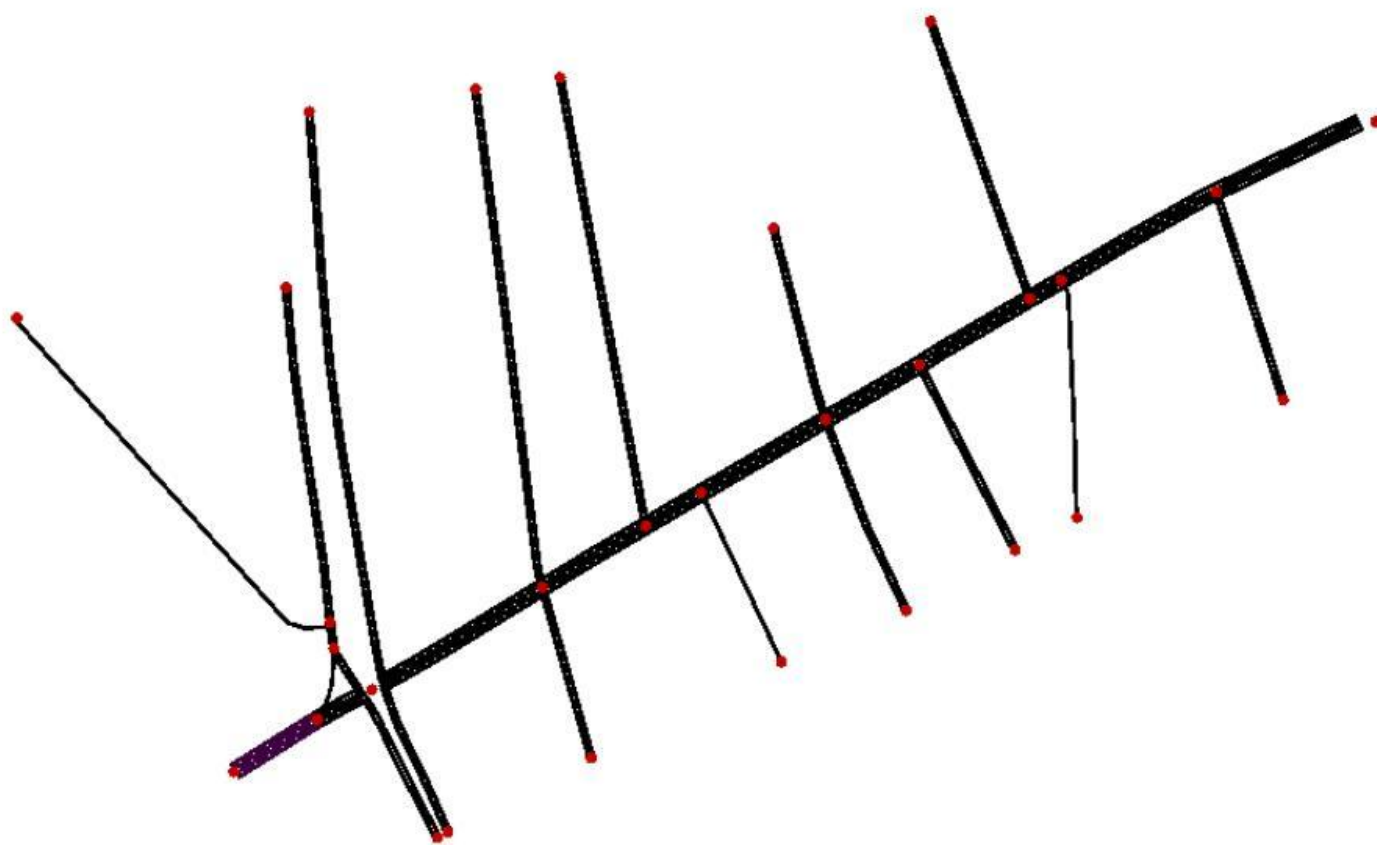}
    \caption{Cologne3}
    \label{fig:cologne3}
  \end{subfigure}
  \hfill
  \begin{subfigure}{0.23\textwidth}
    \includegraphics[width=\linewidth]{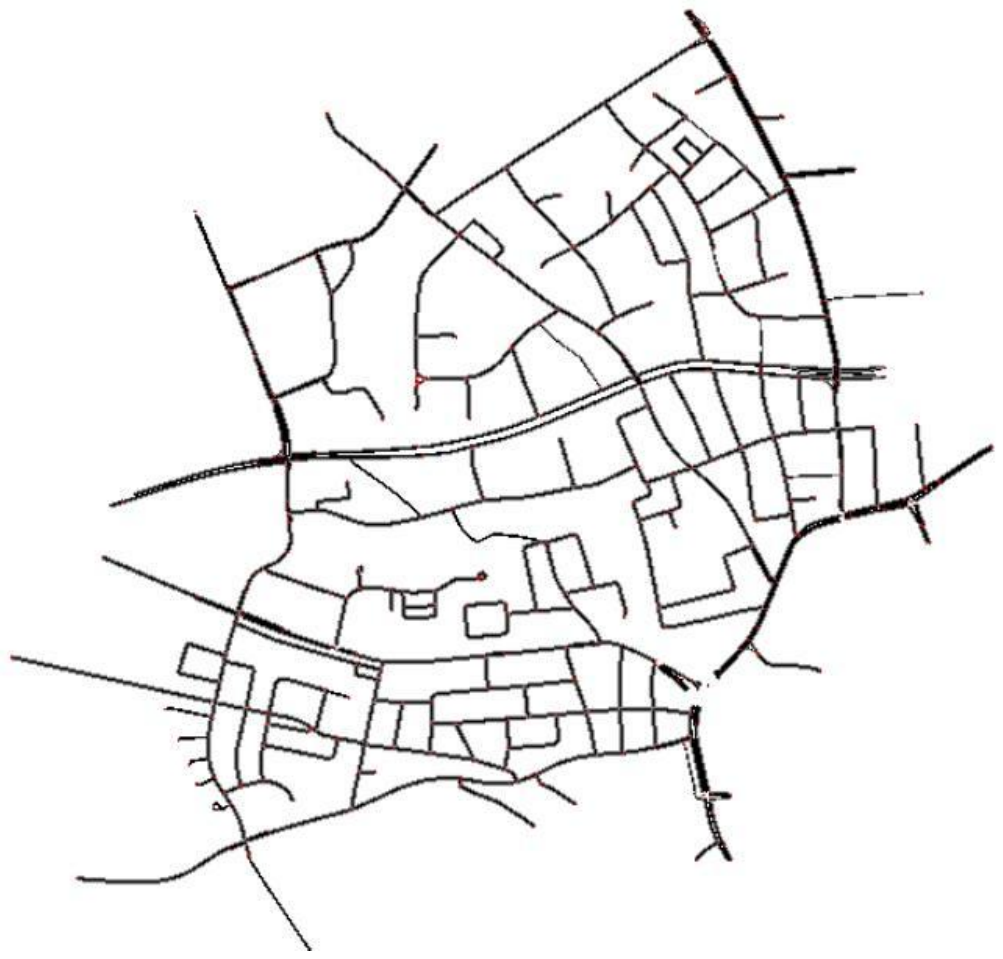}
    \caption{Ingolstadt21}
    \label{fig:ingolstad21}
  \end{subfigure}
  \caption{Real-time road network datasets used in the experiments.}
  \label{fig:real-time-network}
\end{figure}

We conducted experiments on two synthetic datasets and two real-world datasets. The first synthetic dataset \cite{r27} is a custom-designed Manhattan-style traffic network, as shown in Figure \ref{fig:real-time-network}, where each horizontal road is 300 meters long and each vertical road is 200 meters long. All roads are bidirectional with two lanes. Traffic flow was generated using SUMO's \texttt{randomTrips.py}, resulting in a total of 150 vehicles with a vehicle load rate of one vehicle per second. Traffic signals at each intersection were controlled by SUMO's default controller, similar to real-world open-source synthetic datasets.

The second synthetic dataset is also based on a near-realistic open-source synthetic dataset, namely Grid 4x4 \cite{r30}, as shown in Figure \ref{fig:real-time-network}. All roads are 300 meters in length, and there are a total of 16 intersections, each equipped with traffic signal control. We extracted one hour of route data for model training and evaluation, during which 1,473 vehicles entered the road network.

As shown in Figure \ref{fig:real-time-network}, a public real-world dataset, Hangang-4x4, also contains 16 traffic signals \cite{r30}, with each signal controlling one intersection. All W-E (West-East) roads are 800 meters long, and all N-S (North-South) roads are 600 meters long. We extracted 30 minutes of route data from the route files, during which 1,660 vehicles entered the network.

%%%%%%%%%%%%%%%%%%%%%%%%%%%%%%%%%%%%%%
\subsection{Evaluation Metrics}

The evaluation metrics for our task primarily include Average Travel Time, Average Duration, Average Waiting Time, and Average Time Loss. These metrics are defined as follows:
\begin{itemize}
    \item \textbf{Average\_Travel\_Time}: The average time taken by a vehicle to travel from its origin to its destination.
    \item \textbf{Average\_Duration}: The average duration a vehicle spends within the network.
    \item \textbf{Average\_Waiting\_Time}: The average time a vehicle spends waiting due to traffic signals or congestion.
    \item \textbf{Average\_TimeLoss}: The average additional time incurred by a vehicle beyond the ideal travel time, indicating inefficiencies in route planning or traffic management.
\end{itemize}
These metrics are computed based on the outputs recorded from the SUMO simulation software.

%%%%%%%%%%%%%%%%%%%%%%%%%%%%%%%%%%%%%%
\subsection{Comparison with Existing Methods}

We compared the DynamicRouteGPT framework with several representative methods, including:
\begin{itemize}
    \item \textbf{DQNNavigation} \cite{r12}: A multi-agent independent DQN model designed for navigation across multiple intersections, operating in dynamic traffic environments.
    \item \textbf{XRouting} \cite{r27}: An interpretable multi-agent PPO model for navigation tasks, which dynamically learns road and vehicle attributes by adjusting attention and transformer modules to optimize route selection.
    \item \textbf{Dynamic Dijkstra}\cite{r38}: A dynamic path decision-making method that calculates the shortest path at each intersection using the Dijkstra algorithm.
\end{itemize}

%%%%%%%%%%%%%%%%%%%%%%%%%%%%%%%%%%%%%%
\subsection{Parameter Settings}

To evaluate the effectiveness of the DynamicRouteGPT framework, we conducted ablation experiments on three datasets, setting the penetration rate (PR) at 30\%. The penetration rate is defined as follows:
\[
PR = \frac{N_{AV}}{N_{AV} + N_{HV}}
\]
where \(N_{AV}\) represents the number of autonomous vehicles, and \(N_{HV}\) represents the number of human-driven vehicles.

In real-time dynamic traffic route planning, the evaluation metrics can vary even when vehicles follow the same route, due to differences in initial positions, destinations, and departure times. To facilitate a fair comparison, we assigned each vehicle a predefined starting point, destination, and departure time. Additionally, we set the initial speed and acceleration for each vehicle. These controlled settings ensured the validity of the calculated metrics.

%%%%%%%%%%%%%%%%%%%%%%%%%%%%%%%%%%%%%%
\subsection{Performance Evaluation}

Table \ref{tab:performance_comparison} presents the performance of different routing methods across various road networks.

DynamicRouteGPT demonstrates a significant advantage across different testing scenarios. For instance, in the Manhattan dataset, DynamicRouteGPT reduced the average travel time by 31.5\% compared to the XRouting method and by 68.3\% compared to DQN. Additionally, the average waiting time decreased to 37.0 seconds, which is 64.1\% less than the waiting time with XRouting and 81.8\% less than with DQN.

In the Grid 4$\times$4 dataset, DynamicRouteGPT further reduced the average waiting time to 1.67 seconds and the time loss to 7.36 seconds, achieving reductions of 51.5\% and 12.3\%, respectively, compared to XRouting. This result highlights its effectiveness in minimizing traffic delays.

In the more complex Hangzhou 4$\times$4 dataset, DynamicRouteGPT maintained superior performance, reducing the average travel time by 7.7\% compared to DQN and 7.4\% compared to XRouting, and showing a 25.1\% reduction in waiting time compared to dynamic\_DJ.

Overall, DynamicRouteGPT effectively shortens both travel time and waiting time, showcasing its potential as an advanced dynamic navigation framework, particularly suitable for improving the efficiency of urban traffic networks.

\begin{table*}[htbp]
\centering
\caption{Performance Comparison of Different Routing Methods on Various Metrics Across Different Test Scenarios.}
\label{tab:performance_comparison}
\begin{tabular*}{\textwidth}{@{\extracolsep{\fill}} l l c c c c}
    \toprule
    \textbf{File} & \textbf{Routing Method} & \multicolumn{4}{c}{\textbf{Average Metrics}} \\
    \cmidrule(lr){3-6}
    & & \textbf{Travel Time} & \textbf{Duration} & \textbf{Waiting Time} & \textbf{Time Loss} \\
    \midrule
    \multirow{4}{*}{\textbf{Manhattan}} 
    & dynamic\_DJ & 235.1 & 230.0 & 103.0 & 126.62 \\
    & DQN & 507.00 & 506.0 & 203.00 & 279.23 \\
    & Xrouting & 217.00 & 417.0 & 156.00 & 218.63 \\
    & DynamicRouteGPT & 161.00 & 9.0 & 37.00 & 55.36 \\
    \midrule
    \multirow{4}{*}{\textbf{Grid 4$\times$4}} 
    & dynamic\_DJ & 1894.67 & 83.73 & 16.05 & 28.01 \\
    & DQN & 1803.06 & 49.42 & 3.50 & 10.72 \\
    & Xrouting & 1796.80 & 48.55 & 3.44 & 8.39 \\
    & DynamicRouteGPT & 1790.55 & 38.57 & 1.67 & 7.36 \\
    \midrule
    \multirow{4}{*}{\textbf{Hangzhou 4$\times$4}} 
    & dynamic\_DJ & 575.5 & 250.7 & 82.14 & 79.70 \\
    & DQN & 618.20 & 295.5 & 43.60 & 73.11 \\
    & Xrouting & 616.05 & 290.3 & 42.85 & 72.25 \\
    & DynamicRouteGPT & 570.60 & 246.0 & 60.00 & 78.82 \\
    \bottomrule
\end{tabular*}
\end{table*}
%%%%%%%%%%%%%%%%%%%%%%%%%%%%%%%%%%%%%%%%%%%%%%%%%%%%%%%%%
\subsection{Ablation Study}
\begin{figure}[!htb]
  \centering
  \begin{subfigure}[b]{0.15\textwidth}
    \includegraphics[width=\textwidth]{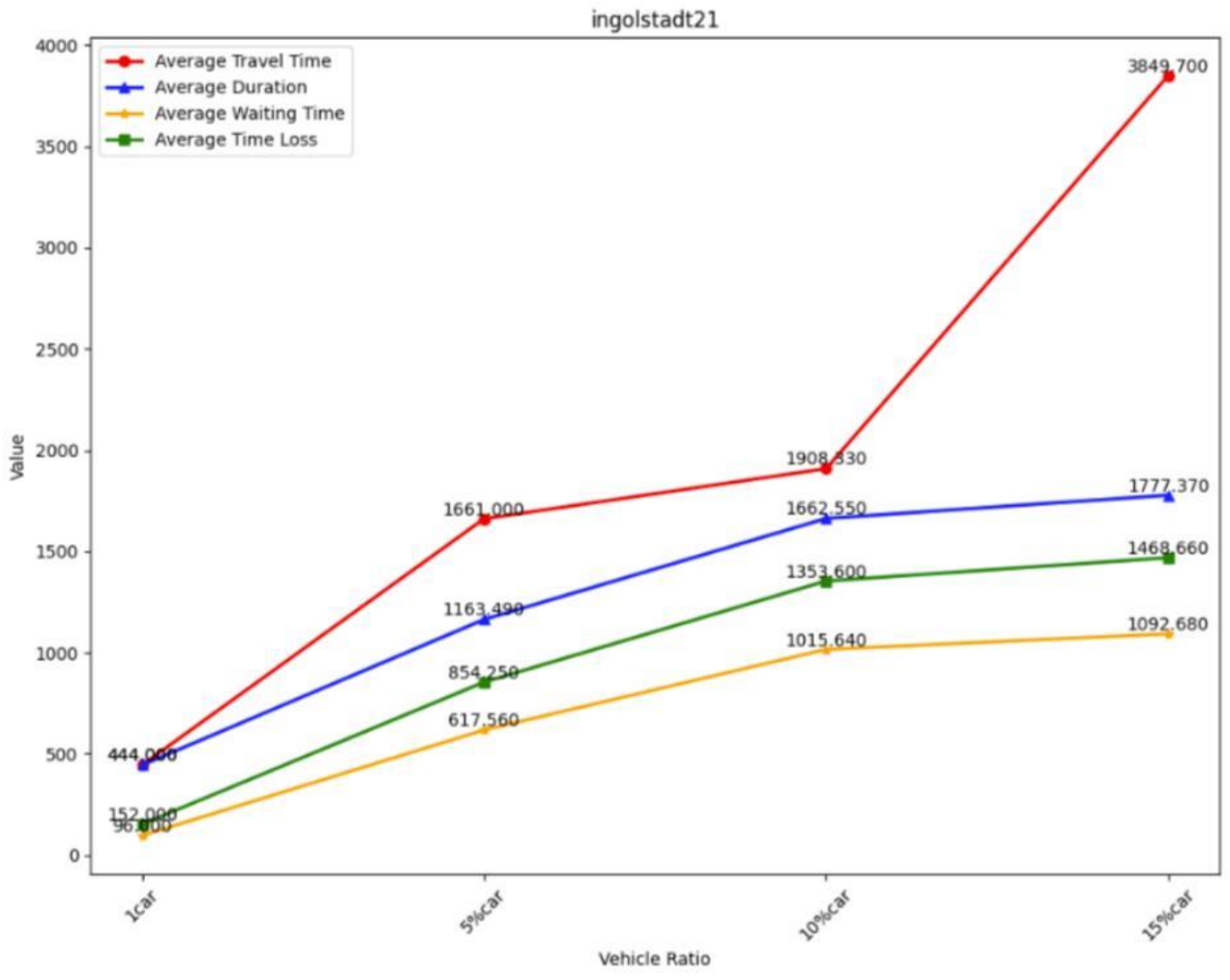}
    \caption{Ingolstadt 21}
    \label{fig:ingolstad_21}
  \end{subfigure}
  \hfill
  \begin{subfigure}[b]{0.15\textwidth}
    \includegraphics[width=\textwidth]{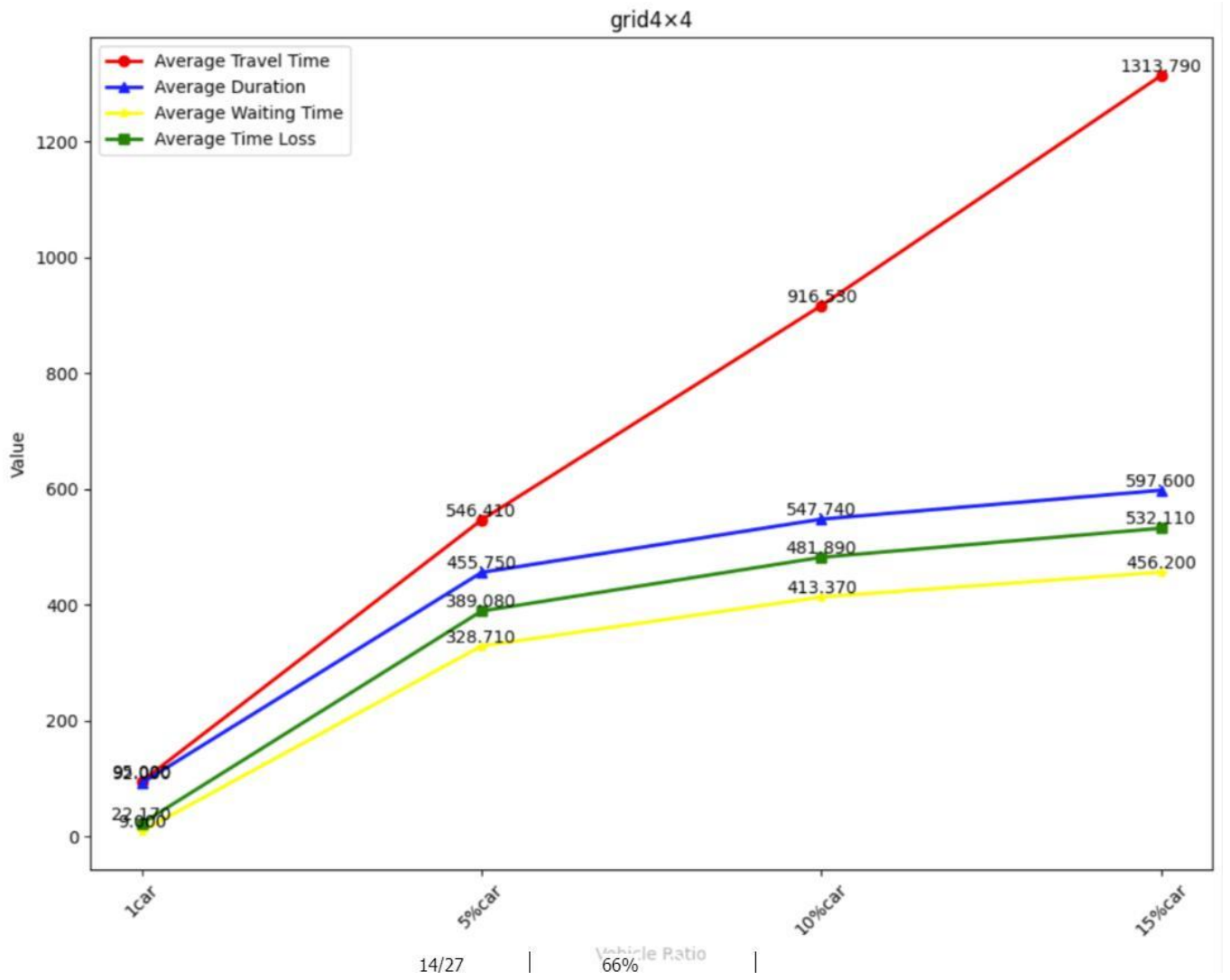}
    \caption{Grid}
    \label{fig:gri_d}
  \end{subfigure}
  \hfill
  \begin{subfigure}[b]{0.15\textwidth}
    \includegraphics[width=\textwidth]{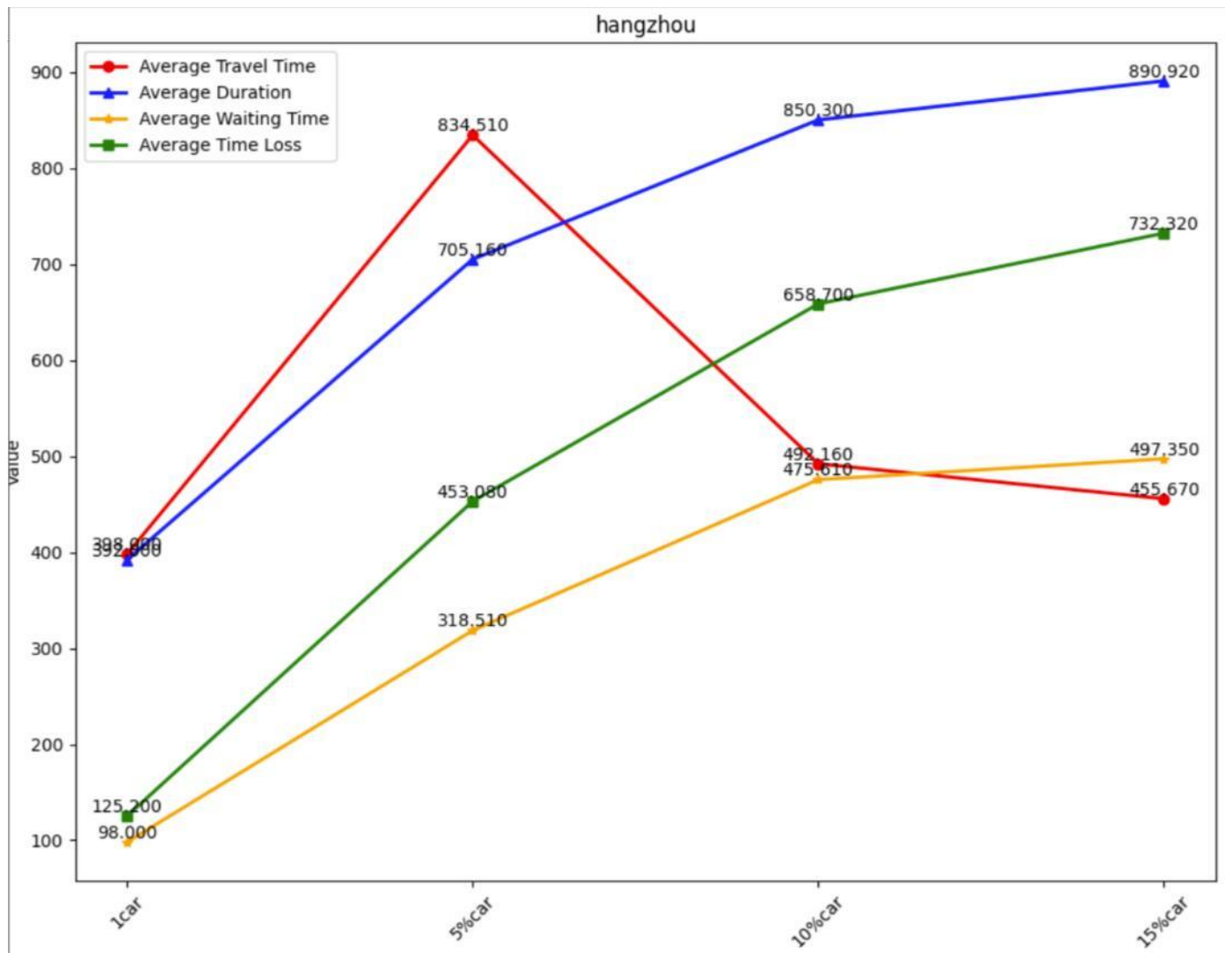}
    \caption{Hangzhou}
    \label{fig:hangzho_u}
  \end{subfigure}
  \hfill
  \begin{subfigure}[b]{0.2\textwidth}
    \includegraphics[width=\textwidth]{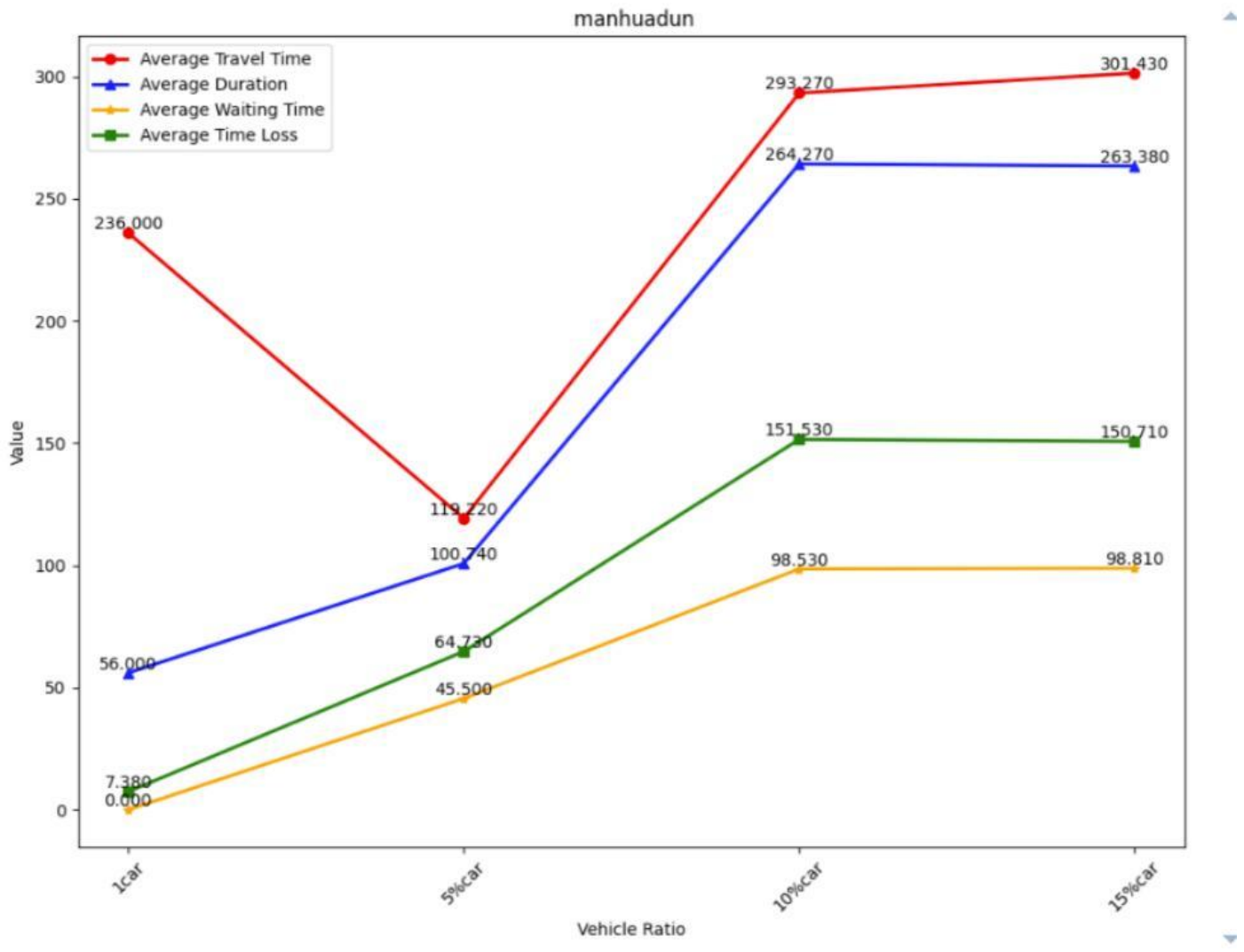}
    \caption{Manhattan}
    \label{fig:Manhatta_n}
  \end{subfigure}
  \hfill
  \begin{subfigure}[b]{0.2\textwidth}
    \includegraphics[width=\textwidth]{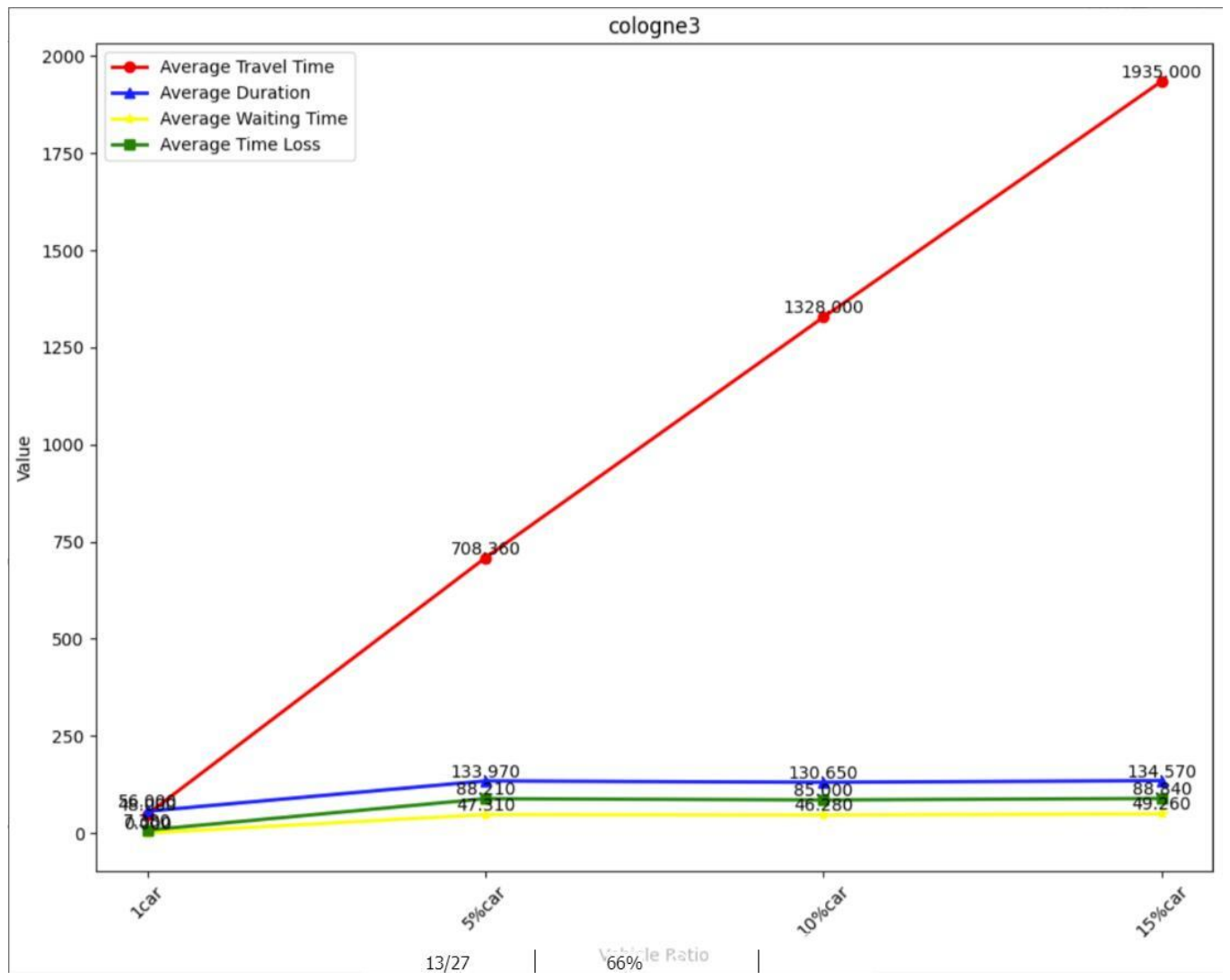}
    \caption{Cologne 3}
    \label{fig:colonge_3}
  \end{subfigure}
  \caption{Real-time road networks for various cities.}
  \label{fig:real-time-networks}
\end{figure}

In addition, for the task of dynamic navigation under real-time road conditions, we observed that different vehicles departing at different times, and from various starting points to different endpoints, experience varying travel times. This variability is particularly influenced by traffic lights and other factors such as waiting times. Based on these observations, we incorporated two additional road network datasets, Ingolstadt21 \cite{r2} and Cologne3 \cite{r2}, as illustrated in Figure \ref{fig:real-time-networks}. Furthermore, we calculated the time required to travel from the same starting point to the same endpoint using the DynamicRouteGPT framework across these different road networks, establishing new baseline benchmarks. These new datasets and benchmarks further validate the performance and adaptability of DynamicRouteGPT in diverse road network environments.

We also analyzed the travel times required using the DynamicRouteGPT framework as the Request Probability (RP) gradually increased to 15\% across different road networks. The results further assess the effectiveness of DynamicRouteGPT, demonstrating its strong generalization capability and suitability for various road networks, thereby confirming the universal applicability of the framework. The framework's performance is depicted in Figure \ref{fig:real-time-networks}.

\begin{table*}[htbp]
\centering
\caption{Performance Comparison on Different Networks with Varying Request Probabilities}
\label{tab:comparison}
\begin{tabular*}{\textwidth}{@{\extracolsep{\fill}} l l l l c c c c}
    \toprule
    \textbf{Network} & \textbf{Start Edge} & \textbf{End Edge} & \textbf{Total Car Num} & \multicolumn{4}{c}{\textbf{Request Probability (\% Cars)}} \\
    \cmidrule(lr){5-8}
    & & & & \textbf{1\%} & \textbf{5\%} & \textbf{10\%} & \textbf{15\%} \\
    \midrule
    Ingolstadt21 & -30482615\#4 & -315358244 & 4283 & 1 & 214 & 428 & 642 \\
    Grid & top2C3 & D3top3 & 1494 & 1 & 75 & 150 & 225 \\
    Hangzhou & road\_0\_4\_0 & road\_1\_2\_2 & 1660 & 1 & 83 & 166 & 249 \\
    Manhattan & right0D0 & A2left2 & 150 & 1 & 7 & 15 & 21 \\
    Cologne3 & 31864804 & -4999334 & 2856 & 1 & 285 & 570 & 855 \\
    \bottomrule
\end{tabular*}
\end{table*}

Table \ref{tab:performance_metrics} presents the evaluation results of DynamicRouteGPT across different road network environments. Specifically, we analyzed the performance of DynamicRouteGPT under varying penetration rates (from 1\% to 15\%). The evaluation metrics include Average Travel Time, Average Duration, Average Waiting Time, and Average Time Loss. These metrics are used to assess the efficiency and stability of the framework under different network conditions.

\begin{table*}[htbp]
    \centering
    \caption{Average Performance Metrics Across Different Road Networks}
    \label{tab:performance_metrics}
    \begin{tabular*}{\textwidth}{@{\extracolsep{\fill}} l S S S S}
        \toprule
        \textbf{Network} & \textbf{Average Travel Time} & \textbf{Average Duration} & \textbf{Average Waiting Time} & \textbf{Average Time Loss} \\
        \midrule
        Ingolstadt21 & 444 & 444 & 96 & 152 \\
        & 1661 & 1163.49 & 617.56 & 854.25 \\
        & 1908.33 & 1662.55 & 1015.64 & 1353.60 \\
        & 3849.70 & 1777.37 & 1092.68 & 1468.66 \\
        \midrule
        Grid 4$\times$4 & 95 & 92 & 9 & 22.17 \\
        & 546.41 & 455.75 & 328.71 & 389.08 \\
        & 916.53 & 547.74 & 413.37 & 481.89 \\
        & 1313.79 & 597.60 & 456.20 & 532.11 \\
        \midrule
        Hangzhou 4$\times$4 & 398 & 392 & 98 & 25.20 \\
        & 834.51 & 705.16 & 318.51 & 453.08 \\
        & 492.16 & 850.30 & 475.61 & 658.70 \\
        & 455.67 & 890.92 & 497.35 & 732.32 \\
        \midrule
        Manhattan & 236 & 56 & 0 & 7.38 \\
        & 119.22 & 100.74 & 45.50 & 64.73 \\
        & 293.27 & 264.27 & 98.53 & 151.53 \\
        & 301.43 & 263.38 & 98.81 & 150.71 \\
        \midrule
        Cologne3 & 48 & 56 & 0 & 7.38 \\
        & 708.36 & 133.97 & 47.31 & 88.21 \\
        & 1328 & 130.65 & 46.28 & 85.00 \\
        & 1935 & 134.57 & 49.26 & 88.84 \\
        \bottomrule
    \end{tabular*}
\end{table*}

The results in Table \ref{tab:performance_metrics} further validate the generalization capability of DynamicRouteGPT across various road network environments. Specifically, DynamicRouteGPT outperforms other baseline methods, whether in the relatively simple Grid network or in more complex and dynamic urban road networks such as Hangzhou and Manhattan.

In these experiments, we gradually increased the penetration rate of autonomous vehicles, reaching up to 15\%. Even at higher penetration rates, DynamicRouteGPT significantly reduced average travel time and waiting time, demonstrating that the framework performs excellently not only under low-density traffic conditions but also maintains superior performance in the face of higher-density, more complex traffic flows.

More specifically, in the simpler Grid 4×4 network, DynamicRouteGPT effectively reduced the average travel time and waiting time, indicating its highly efficient path-planning capability in relatively regular network topologies. In the more complex Hangzhou 4×4 network, despite increased traffic flow and more complicated road structures, DynamicRouteGPT still exhibited strong adaptability, intelligently responding to intricate road network structures to optimize path selection and thus reduce travel and waiting times. Similarly, in the Manhattan urban road network, DynamicRouteGPT demonstrated robust performance, successfully addressing the challenges posed by diverse road segments and high traffic volumes.

\begin{figure}[!ht]
  \centerline{\includegraphics[width=1\linewidth]{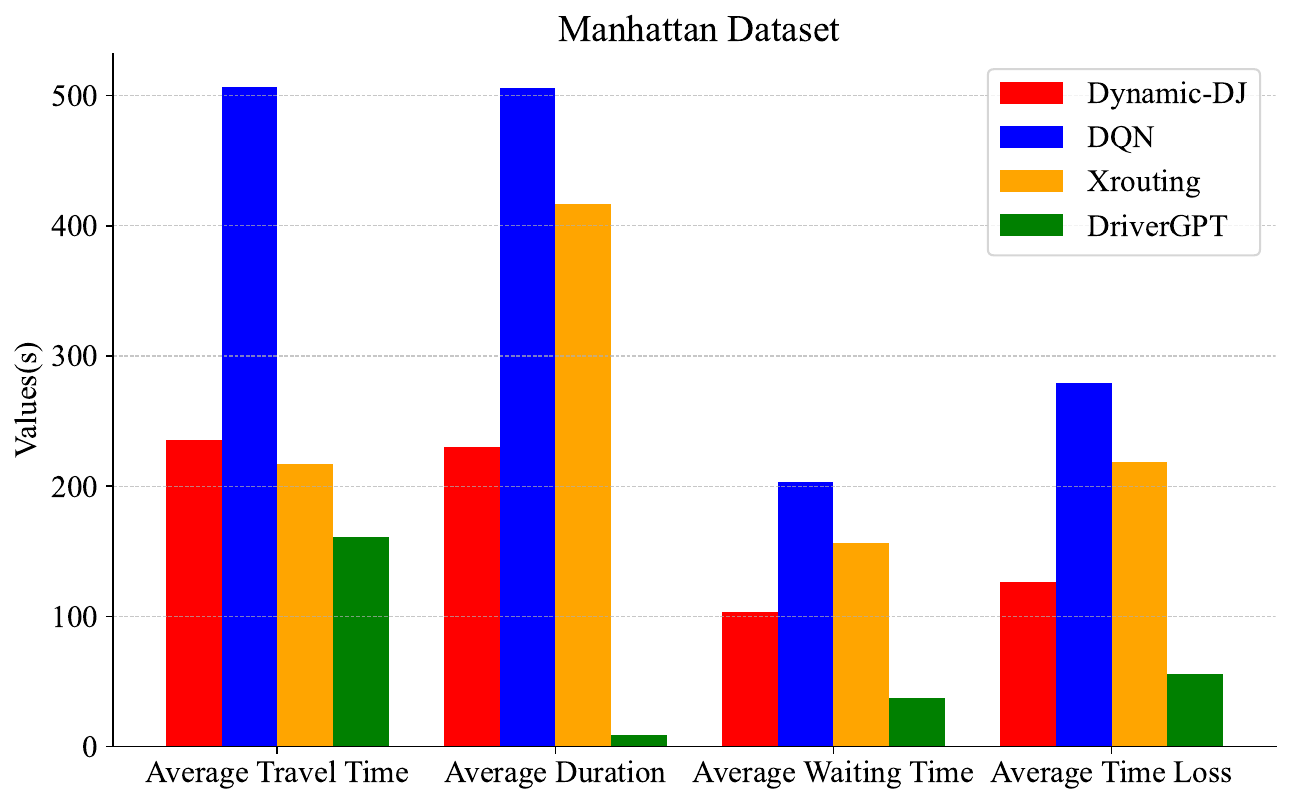}}
  \caption{Performance Comparison on Manhattan}
  \label{Manhattan}
\end{figure}

\begin{figure}[!t]
  \centerline{\includegraphics[width=1\linewidth]{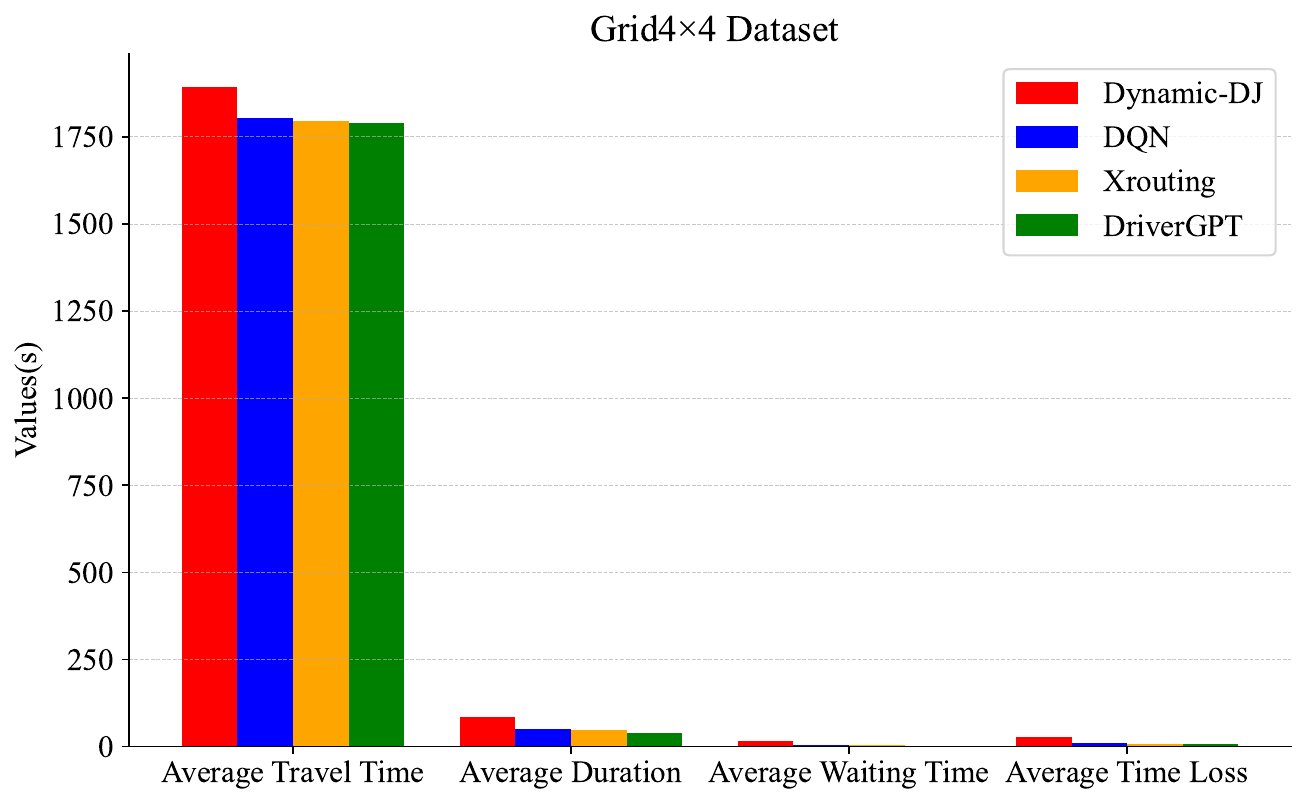}}
  \caption{Performance Comparison on grid}
  \label{grid4}
\end{figure}

\begin{figure}[!t]
  \centerline{\includegraphics[width=1\linewidth]{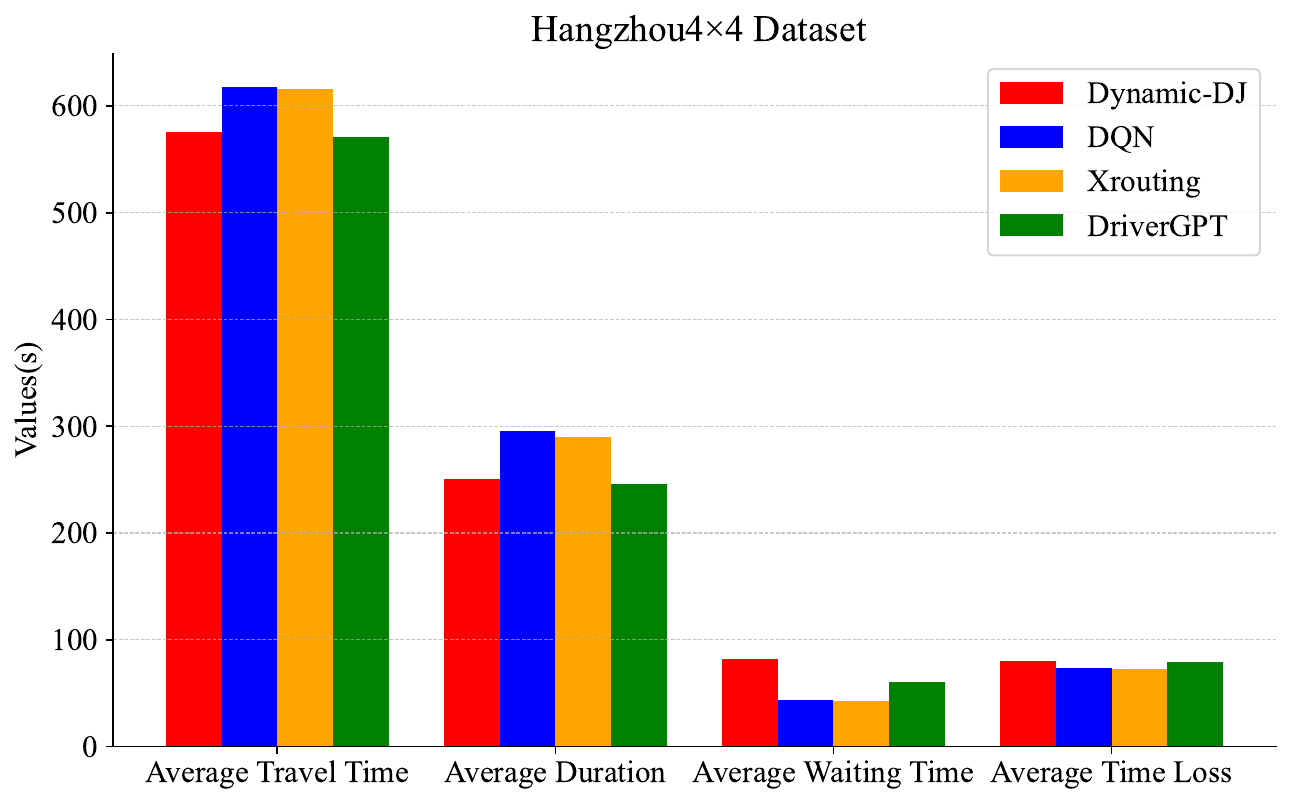}}
  \caption{Performance Comparison on hangzhou}
  \label{hangzhou4}
\end{figure}

Overall, these results demonstrate that DynamicRouteGPT not only adapts well to various road network structures but also maintains stable performance under different traffic conditions. This strong generalization ability makes the DynamicRouteGPT framework not only suitable for specific scenarios but also broadly applicable across various complex urban traffic networks, providing efficient and reliable dynamic route planning services.These results collectively showcase the potential of DynamicRouteGPT as an advanced dynamic navigation framework, particularly suited for enhancing the efficiency of urban traffic networks. The performance comparison is more intuitively illustrated in Figures \ref{Manhattan}, \ref{grid4}, and \ref{hangzhou4}.

%%%%%%%%%%%%%%%%%%%%%%%%%%%%%%%%%%%%%%
\subsection{Parameter Analysis}

We conducted a sensitivity analysis of the key parameters in the DynamicRouteGPT framework to evaluate its performance under different settings. The experimental results demonstrate that DynamicRouteGPT exhibits strong robustness and adaptability across various traffic conditions and vehicle penetration rates. By appropriately adjusting these parameters, the framework's performance can be further optimized to better suit specific traffic environments, thereby enhancing the efficiency and accuracy of route planning.

To further validate the practical application of the DynamicRouteGPT framework, we performed several case studies. These case studies illustrate the path selection process and outcomes of the framework under different scenarios. Detailed information can be found in Appendix A.

%%%%%%%%%%%%%%%%%%%%%%%%%%%%%%%%%%%%%%
\subsection{Limitations Analysis}

Despite the superior performance of DynamicRouteGPT demonstrated in multiple experiments, there are certain limitations associated with the framework. First, the computational complexity of the model is relatively high, which may pose challenges in large-scale real-time applications due to potential computational resource constraints. Second, although DynamicRouteGPT is capable of adapting to various road network environments, further optimization may be required in extremely large-scale, high-density urban road networks to enhance its efficiency. Additionally, the effectiveness of the model heavily relies on the accuracy and completeness of real-time traffic information; any inaccuracies in the input data could adversely affect the accuracy of route planning.

In conclusion, while the DynamicRouteGPT framework exhibits strong performance and broad applicability in dynamic route planning tasks, future research should focus on further optimization and expansion to address more complex traffic scenarios and a wider range of application demands.

%%%%%%%%%%%%%%%%%%%%%%%%%%%%%%%%%%%%%%%%%%%%%%%%%%%%%%%%%%%%%%%%%%%%%%%%%
\section{Conclusion}

This study presented and validated the DynamicRouteGPT framework, a novel approach to dynamic route planning in complex traffic scenarios. Through extensive experiments, the framework demonstrated consistent superiority over traditional methods, particularly in reducing average travel time, waiting time, and time loss across diverse road networks.

Unlike many existing models, DynamicRouteGPT does not require pre-training, which significantly reduces computational overhead while enhancing adaptability in real-world conditions. The framework’s strong generalization ability allows it to efficiently manage various traffic conditions and vehicle penetration rates, further optimized through parameter fine-tuning to meet specific traffic environment needs.

A key contribution of this work is the establishment of a new baseline for evaluating dynamic path planning under real-time traffic conditions. This baseline provides a useful reference for assessing the performance of path planning algorithms across different road networks, reinforcing the practical applicability of DynamicRouteGPT.

Case studies confirmed the framework's effectiveness in managing complex scenarios, including standard path selection, forced routing, and emergency response. These findings offer valuable insights for improving urban traffic management systems, with potential benefits in enhancing traffic flow efficiency and safety.

While the DynamicRouteGPT framework exhibits several advantages, its computational complexity remains a challenge, particularly for large-scale real-time applications. Additionally, the accuracy of the model’s predictions is dependent on the quality of real-time traffic data, highlighting the need for further optimization in cases of incomplete or noisy data.

Future work will explore the integration of large language models with traffic signal control systems, aiming to optimize vehicle routing and traffic signal operations simultaneously. Moreover, investigating alternative algorithms, such as suboptimal path selection, could provide more balanced routing options, considering factors like scenic routes or fuel efficiency.

In summary, DynamicRouteGPT represents a significant advancement in dynamic route planning, offering a flexible and efficient solution for intelligent transportation systems. Future research will continue to refine this framework, addressing its limitations and exploring its potential in large-scale applications.

%%%%%%%%%%%%%%%%%%%%%%%%%%%%%%%%%%%%%%%%%%%%%%%%%%%%%%%%%%%%%%%%%%%%%%%%%
\bibliographystyle{ACM-Reference-Format}

%%%%%%%%%%%%%%%%%%%%%%%%%%%%%%%%%%%%%%%%%%%%%%%%%%%%%%%%%%%%%%%%%%%%%%%
\appendix
\section{Appendix Title 1}
\subsection{Standard Route Selection}

Vehicle \texttt{veh\_42} has arrived at intersection \texttt{A2}. The current properties of intersection \texttt{A2} are \texttt{traffic\_light\_right\_on\_red}, and the node ID is \texttt{A2}, with the target edge being \texttt{B2C2}.

There are several alternative paths available:
\begin{itemize}
    \item Alternative path 1: [\texttt{A3A2}, \texttt{A2B2}, \texttt{B2C2}]
    \item Alternative path 2: [\texttt{A3A2}, \texttt{A2A1}, \texttt{A1B1}, \texttt{B1B2}, \texttt{B2C2}]
    \item Alternative path 3: [\texttt{A3A2}, \texttt{A2A1}, \texttt{A1AO}, \texttt{ABBe}, \texttt{BOB1}, \texttt{B1B2}, \texttt{B2C2}]
\end{itemize}

The chosen path is: [\texttt{A3A2}, \texttt{A2B2}, \texttt{B2C2}]

The choice reason is as follows:
\begin{align*}
    \text{Total time: } & \text{path1} < \text{path2} < \text{path3} \\
    \text{Traffic light count: } & \text{path1} < \text{path2} < \text{path3} \\
    \text{Error edges: } & \text{none}
\end{align*}

Taking all factors into consideration, path1 is selected.

The best route for \texttt{veh\_42} is: [\texttt{top1B3}, \texttt{A3A2}, \texttt{A2B2}, \texttt{B2C2}, \texttt{C2D2}, \texttt{D2right2}].
%%%%%%%%%%%%%%%%%%%%%%%%%%%%%%%%%%%%%%%%%%%%%%%%%%%%%%%%%%%%%%%%%%%%%%%%%%%%%%%%%%%%%%%%%%%%%%%%%%%%

\subsection{Adding A Mandatory Route}

Minimizing overall travel time while ensuring the route passes through a specified path.

Vehicle \texttt{veh\_42} has arrived at intersection \texttt{A2}. The current properties of intersection \texttt{A2} are \texttt{traffic\_light\_right\_on\_red}, and the node ID is \texttt{A2}, with the target edge being \texttt{B2C2}.

There are several alternative paths available:
\begin{itemize}
    \item Alternative path 1: [\texttt{A3A2}, \texttt{A2B2}, \texttt{B2C2}]
    \item Alternative path 2: [\texttt{A3A2}, \texttt{A2A1}, \texttt{A1B1}, \texttt{B1B2}, \texttt{B2C2}]
    \item Alternative path 3: [\texttt{A3A2}, \texttt{A2A1}, \texttt{A1AO}, \texttt{ABBe}, \texttt{BOB1}, \texttt{B1B2}, \texttt{B2C2}]
\end{itemize}

The chosen path is: [\texttt{A3A2}, \texttt{A2A1}, \texttt{A1B1}, \texttt{B1B2}, \texttt{B2C2}]

The choice reason is as follows:
\begin{align*}
    \text{Total time: } & \text{path1} < \text{path2} < \text{path3} \\
    \text{Traffic light count: } & \text{path1} < \text{path2} < \text{path3} \\
    \text{Error edges: } & \text{none} \\
    \text{Mandatory path: } & \text{\texttt{A2A1}}
\end{align*}

Taking all factors into consideration, path2 is selected.

The best route for \texttt{veh\_42} is: [\texttt{top1B3}, \texttt{A3A2}, \texttt{A2A1}, \texttt{A1B1}, \texttt{B1B2}, \texttt{B2C2}, \texttt{C2D2}, \texttt{D2right2}].

%%%%%%%%%%%%%%%%%%%%%%%%%%%%%%%%%%%%%%%%%%%%%%%%%%%%%%%%%%%%%%%%%%%%%%%%%%%%%%%%%%%%%%%%%%%%%%%%%%%%
\subsection{Handling Emergencies}

Minimizing overall travel time under the constraint of avoiding a specific path.

Vehicle \texttt{veh\_42} has arrived at intersection \texttt{A2}. The current properties of intersection \texttt{A2} are \texttt{traffic\_light\_right\_on\_red}, and the node ID is \texttt{A2}, with the target edge being \texttt{B2C2}.

There are several alternative paths available:
\begin{itemize}
    \item Alternative path 1: [\texttt{A3A2}, \texttt{A2B2}, \texttt{B2C2}]
    \item Alternative path 2: [\texttt{A3A2}, \texttt{A2A1}, \texttt{A1B1}, \texttt{B1B2}, \texttt{B2C2}]
    \item Alternative path 3: [\texttt{A3A2}, \texttt{A2A1}, \texttt{A1AO}, \texttt{ABBe}, \texttt{BOB1}, \texttt{B1B2}, \texttt{B2C2}]
\end{itemize}

The chosen path is: [\texttt{A3A2}, \texttt{A2A1}, \texttt{A1AO}, \texttt{ABBe}, \texttt{BOB1}, \texttt{B1B2}, \texttt{B2C2}]

The choice reason is as follows:
\begin{align*}
    \text{Total time: } & \text{path1} < \text{path2} < \text{path3} \\
    \text{Traffic light count: } & \text{path1} < \text{path2} < \text{path3} \\
    \text{Error edges: } & \text{\texttt{A1B1}} \\
    \text{Mandatory path: } & \text{\texttt{A2A1}}
\end{align*}

Taking all factors into consideration, path3 is selected.

The best route for \texttt{veh\_42} is: [\texttt{top1B3}, \texttt{A3A2}, \texttt{A2A1}, \texttt{A1AO}, \texttt{ABBe}, \texttt{BOB1}, \texttt{B1B2}, \texttt{B2C2}, \texttt{C2D2}, \texttt{D2right2}].
\end{document}